\pdfoutput=1
\documentclass[11pt]{article}

\usepackage{acl}

\usepackage{times}
\usepackage{latexsym}

\usepackage[T1]{fontenc}
\usepackage[utf8]{inputenc}

\usepackage{microtype}

\usepackage{graphicx}
\usepackage{tikz}
\usetikzlibrary{calc}
\usetikzlibrary{positioning}
\usepackage{amssymb,amsmath,amsthm,amsfonts}
\usepackage{enumitem}
\usepackage{booktabs}
\usepackage{multirow}
\usepackage{cleveref}
\usepackage[ruled,linesnumbered,noend]{algorithm2e}
\usepackage{soul}
\DeclareMathOperator*{\softmax}{softmax}

\setlist{noitemsep}

\crefname{algocf}{alg.}{algs.}
\Crefname{algocf}{Algorithm}{Algorithms}

\newcommand{\CLS}{\texttt{\small [CLS]}}
\newcommand{\SEP}{\texttt{\small [SEP]}}

\title{Improving Toponym Resolution with Better Candidate Generation, Transformer-based Reranking, and Two-Stage Resolution}

\author{Zeyu Zhang \and  Steven Bethard\\
  School of Information, The University of Arizona, Tucson, AZ, USA \\
  \texttt{\{zeyuzhang, bethard\}@arizona.edu} \\}
\SetAlFnt{\small}
\begin{document}
\maketitle
\begin{abstract}
Geocoding is the task of converting location mentions in text into structured data that encodes the geospatial semantics.
We propose a new architecture for geocoding, GeoNorm.
GeoNorm first uses information retrieval techniques to generate a list of candidate entries from the geospatial ontology.
Then it reranks the candidate entries using a transformer-based neural network that incorporates information from the ontology such as the entry's population.
This generate-and-rerank process is applied twice: first to resolve the less ambiguous countries, states, and counties, and second to resolve the remaining location mentions, using the identified countries, states, and counties as context.
Our proposed toponym resolution framework achieves state-of-the-art performance on multiple datasets.
Code and models are available at \url{https://github.com/clulab/geonorm}.
\end{abstract}

\section{Introduction}
Geospatial information extraction is a type of semantic extraction that plays a critical role in tasks such as
geographical document classification and retrieval \cite{bhargava-etal-2017-lithium},
historical event analysis based on location data \cite{tateosian2017tracking},
tracking the evolution and emergence of infectious diseases \cite{hay2013global},
and disaster response mechanisms \citep{ashktorab2014tweedr,de2018taggs}.
Such information extraction can be challenging because different geographical locations can be referred to by the same place name (e.g., \textit{San Jose} in Costa Rica vs. \textit{San Jose} in California, USA), and different place names  can refer to the same geographical location (e.g., \textit{Leeuwarden} and \textit{Ljouwert} are two names for the same city in the Netherlands).
It is thus critical to resolve these place names by linking them with their corresponding coordinates from a geospatial ontology or knowledge base.

Geocoding, also called toponym resolution or toponym disambiguation, is the subtask of geoparsing that disambiguates place names (known as \textit{toponyms}) in text.
Given a textual mention of a location, a geocoder chooses the corresponding geospatial coordinates, geospatial polygon, or entry in a geospatial database.
Approaches to geocoding include generate-and-rank systems that first use information retrieval systems to generate candidate entries and then rerank them with hand-engineered heuristics and/or supervised classifiers \citep[e.g.,][]{grover2010use,speriosu-baldridge-2013-text,wang-etal-2019-dm}, vector-space systems that use deep neural networks to encode place names and database entries as vectors and measure their similarity \citep[e.g.,][]{hosseini-etal-2020-deezymatch,ardanuy2020deep}, and tile-classification systems that use deep neural networks to directly predict small tiles of the map rather than ontology entries \citep[e.g.,][]{gritta-etal-2018-melbourne, cardoso2019using,kulkarni-etal-2021-multi}.
The deep neural network tile-classification approaches have been the most successful, but they do not naturally produce an ontology entry, which contains semantic metadata needed by users.

\begin{figure*}
\centering
\begin{tikzpicture}[
  model/.style={draw, fill=black!25, align=center},
  geoentry/.style={draw, fill=white, inner sep=1pt, minimum height=6ex, text width=0.29\textwidth},
  textforentry/.style={draw, fill=white, text width=0.28\textwidth, align=left},
  font=\scriptsize,
]

\node[draw, text width=\textwidth, align=left] (text) at (0, 0) {\textbf{Alberta}'s capital city sits in eighth place out of 10 Canadian cities for its socio-economic and physical health \ldots for whatever reason, is quite high in \textbf{Edmonton} compared to other cities \ldots The Conference Board of \textbf{Canada} cautioned that benchmarking is not an end onto itself\ldots};

\node[model, text width=\textwidth, below=.8cm of text] (geonames) {GeoNames index (Lucene)};

\path
  let \p1 = (text.south) in
  let \p2 = (geonames.north) in
  node[inner sep=1pt] (mention1) at (-5.5, -.8) {Alberta} edge[latex-] (-5.5, \y1) edge[-latex] (-5.5, \y2)
  node[inner sep=1pt] (mention2) at (0, -.8) {Edmonton} edge[latex-] (0, \y1) edge[-latex] (0, \y2)
  node[inner sep=1pt] (mention3) at (5.5, -.8) {Canada} edge[latex-] (5.5, \y1) edge[-latex] (5.5, \y2);

\foreach[count=\i] \solution/\entries in {
    \includegraphics[scale=.4]{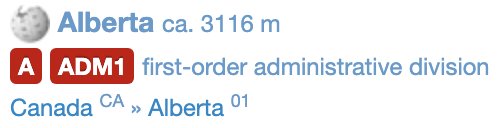}/{
      2/
      \includegraphics[scale=.4]{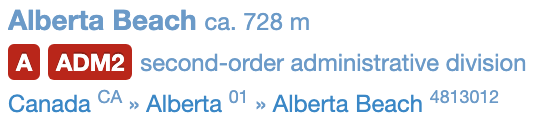}/
      [CLS] Alberta [SEP] Alberta Beach [SEP] \ldots/,
      1/
      \includegraphics[scale=.4]{geonames/Alberta1.png}/
      [CLS] Alberta [SEP] Alberta [SEP] AB [SEP] Alb. \ldots/},
    \includegraphics[scale=.4]{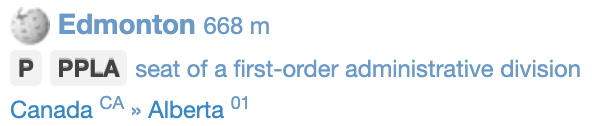}/{
      2/
      \includegraphics[scale=.4]{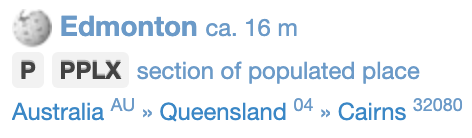}/
      [CLS] Edmonton [SEP] Edmonton | US | KY [SEP] Ehdmonton\ldots/
      [CLS] Edmonton | CA | 01 [SEP] Edmonton | US | KY [SEP] Ehdmonton\ldots,
      1/
      \includegraphics[scale=.4]{geonames/Edmonton1.png}/
      [CLS] Edmonton [SEP] Edmonton [SEP] Edmontona\ldots/
      [CLS] Edmonton | CA | 01 [SEP] Edmonton | CA | 01 [SEP] Edmontona\ldots},
    \includegraphics[scale=.4]{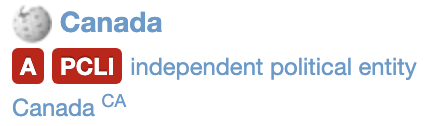}/{
      2/
      \includegraphics[scale=.4]{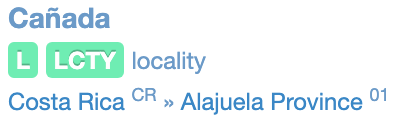}/
      [CLS] Canada [SEP] Canada [SEP] Cañada\ldots/,
      1/
      \includegraphics[scale=.4]{geonames/Canada1.png}/
      [CLS] Canada [SEP] Canada [SEP] Kanadaa\ldots/}} {

  % softmax
  \node[model, text width=.3\textwidth, below=6cm of mention\i] (softmax\i) {Softmax};

  % solutions
  \ifthenelse{\i=2}{
    % repeated softmax
    \node[model, text width=.3\textwidth, below=3.3cm of softmax\i] (resoftmax\i) {Softmax};
    \node[geoentry, below=.4cm of resoftmax\i] (solved\i) {\solution} edge[latex-] (resoftmax\i);
  }{
    \node[geoentry, below=.4cm of softmax\i] (solved\i) {\solution} edge[latex-] (softmax\i);
  }

  \foreach[evaluate=\j as \xshift using .5*(\j-1)-.25,
           evaluate=\j as \yshift using .25*(\j-1)]
      \j/\entry/\text/\retext in \entries {
    % GeoNames entries
    \path
      let \p1 = (mention\i.south) in
      let \p2 = (geonames.south) in
      node[geoentry, below=1.4cm of mention\i, shift={(\xshift, \yshift)}] (entry\i\j) {\entry \hfill $e_{\i\j}$}
      edge[latex-] (\x1, \y2);

    % transformer input text
    \node[textforentry, below=.6cm of entry\i\j] (transformerin\i\j) {\text}
    edge[latex-] (entry\i\j.south);

    % softmax input features
    \node[textforentry, below=1.3cm of transformerin\i\j] (softmaxin\i\j) {$h_{\textsc{[CLS]}} \oplus \log(\textsc{Pop}(e_{\i\j})) \oplus \textsc{Type}(e_{\i\j})$}
    edge [latex-] (transformerin\i\j.south) edge[-latex] (softmax\i);

    \ifthenelse{\i=2}{
      % transformer input text
      \node[textforentry, below=3.2cm of transformerin\i\j] (retransformerin\i\j) {\retext};

      % softmax input features
      \node[textforentry, below=1.1cm of retransformerin\i\j] (resoftmaxin\i\j) {$h_{\textsc{[CLS]}} \oplus \log(\textsc{Pop}(e_{\i\j})) \oplus \textsc{Type}(e_{\i\j})$}
      edge[latex-] (retransformerin\i\j.south) edge[-latex] (resoftmax\i);
    }{}
  }
}

% transformers (here so they're on top of the arrows)
\node[model, text width=\textwidth, below=.35cm of transformerin21, shift={(.25, 0)}] (transformer) {Transformer};
\node[model, text width=.3\textwidth, below=1.5cm of softmax2] (retransformer) {Transformer};

% special edges
\foreach \j in {1,2} {
  \draw[-latex] (transformerin2\j.south) to[bend left] (retransformerin2\j);
  \draw[-latex] (solved1.east) to[bend left] (retransformerin2\j);
  \draw[-latex] (solved3.west) to[bend right] (retransformerin2\j);
}

\end{tikzpicture}
\caption{The architecture of our model, GeoNorm, applied to a sample text. The location mentions to be resolved are in bold.}
\label{fig:architecture}
\end{figure*}

We propose a new architecture, GeoNorm, shown in \Cref{fig:architecture}, which builds on all of these lines of research:
it uses pre-trained deep neural networks for the improved robustness in matching place names, while leveraging a generate-then-rank architecture to produce ontology entries as output.
It couples this generate-and-rank process with a two-stage approach that first resolves the less ambiguous countries, states, and counties, and then resolves the remaining location mentions, using the identified countries, states, and counties as context.

Our work makes the following contributions:
\begin{itemize}
\item Our proposed architecture for geocoding achieves new state-of-the-art performance, outperforming prior work by large margins on toponym resolution corpora: 19.6\% improvement on Local Global Lexicon (LGL), 9.0\% on GeoWebNews, and 16.8\% on TR-News.
\item Our candidate generator alone, based on simple information retrieval techniques, outperforms more complex neural models, demonstrating the importance of establishing strong baselines for evaluation.
\item Our reranker is the first application of pre-trained transformers for encoding location mentions and context for toponym resolution.
\item Our two-stage resolution provides a simple and effective new approach to incorporating document-level context for geocoding.
\end{itemize}

\section{Related Work}
The current work focuses on mention-level geocoding.
Related tasks include document-level geocoding and geotagging.
Document-level geocoding takes as input an entire text and produces as output a location from a geospatial ontology, as in geolocating Twitter users or microblog posts \citep{roller-etal-2012-supervised,rahimi-etal-2015-exploiting,lee2015read,rahimi-etal-2017-continuous,hoang2018location,kumar2019location,luo2020overview} and geographic document retrieval and classification \citep{gey2005geoclef,adams2018crowdsourcing}.
Geotagging takes as input an entire text and produces as output a list of location phrases \citep{Gritta2018}.
Mention-level geocoding, the focus of the current article, takes as input location phrases from a text and produces as output their corresponding locations in a geospatial ontology.
This is related to the task of linking phrases to Wikipedia, though geospatial ontologies do not have full text articles for each of their concepts, which are required for training many recent Wikipedia linking approaches \cite[e.g., ][]{yamada-etal-2022-global,ayoola-etal-2022-refined}.

Early systems for mention-level geocoding used hand-crafted rules and heuristics to predict geospatial labels for place names: Edinburgh geoparser \citep{grover2010use}, \citet{tobin2010evaluation}, \citet{lieberman2010geotagging}, \citet{lieberman2011multifaceted}, CLAVIN \citep{clavin2012}, GeoTxt \citep{karimzadeh2013geotxt}, and \citet{laparra-bethard-2020-dataset}.
The most common features and heuristics were based on string matching, population count, and type of place (city, country, etc.).
%
%String matching can be done exactly, or approximately with edit distances metrics like Levenshtein Distance.

Later geocoding systems used heuristics of rule-based systems as features in supervised machine learning models, including logistic regression 
\citep[WISTR,][]{speriosu-baldridge-2013-text},
support vector machines (%
\citealp{martins2010machine};
\citealp{zhang2014geocoding}%
),
random forests (%
\citealp[MG,][]{freire2011metadata};
\citealp{lieberman2012adaptive}%
), stacked LightGBMs \citep[DM\_NLP,][]{wang-etal-2019-dm} and other statistical learning methods (%
\citealp[Topocluster,][]{delozier2015gazetteer};
\citealp[CBH, SHS,][]{kamalloo2018coherent}).
These systems typically applied a generate-then-rerank framework: the mention text is used to query an information retrieval index of the geospatial ontology and produce candidate ontology entries, then a supervised machine-learning model reranks the candidates using additional features.

Some deep learning models approach geocoding as a vector-space problem.
Both the mention text and ontology entries are converted into vectors, and vector similarity is used to select the most appropriate ontology entry for each mention \citep{hosseini-etal-2020-deezymatch,ardanuy2020deep}.
Such approaches should allow more flexible matching of mentions to concepts, but we find that simple information retrieval techniques outperform these models.

Other deep learning models approach geocoding as a classification problem by dividing the Earth's surface into an $N \times N$ grid of tiles.
Place names and their features are mapped to one of these tiles using convolutional (%
\citealp[CamCoder,][]{gritta-etal-2018-melbourne};
\citealp[MLG,][]{kulkarni-etal-2021-multi}%
) or recurrent neural networks (\citealp{cardoso2019using}).
Such approaches can flexibly match mentions to concepts and can also incorporate textual context, but do not naturally produce ontology entries, which contain semantic metadata needed by users.

Our proposed approach combines the tight ontology integration of the generate-and-rerank systems with the robust text and context encoding of the deep neural network classifiers.

\section{Proposed Methods}

We define the task of toponym resolution as follows.
We are given an ontology or knowledge base with a set of entries $E = \{e_1, e_2, ..., e_{|E|}\}$.
Each input is a text made up of sentences $T = \{t_1, t_2, \ldots, t_{|T|}\}$ and a list of location mentions $M = \{m_1, m_2, ..., m_{|M|}\}$ in the text.
The goal is to find a mapping function $f(m_i)=e_{j}$ that maps each  location mention in the text to its corresponding entry in the ontology. 

We approach toponym resolution using a candidate generator followed by a candidate reranker.
The candidate generator, $G(m,E) \to E_m$, takes a mention $m$ and ontology $E$ as input, and generates a list of candidate entries $E_m$, where $E_m \subseteq E$ and $|E_m| \ll |E|$.
As the candidate generator must search a large ontology and produce only a short list of candidates, the goal for $G$ will be high recall and high runtime efficiency.
The candidate reranker, $R(m, E_m) \to \widehat{E_m}$, takes a mention $m$ and the list of candidate ontology entries $E_m$, and sorts them by their relevance or importance to produce a new list, $\widehat{E_m}$.
As the candidate ranker needs to work only with a short list of candidates, the goal for $R$ will be high precision, especially at rank 1, with less of a focus on runtime efficiency.

\subsection{Candidate Generator}
\label{section:generator}

Our candidate generator is inspired by prior work on geocoding in using information retrieval techniques to search for candidates in the ontology \cite{grover2010use, clavin2012}.
Accurate candidate generation is essential, since the generator's recall is the ceiling performance for the reranker.
As we will see in \cref{sec:results}, our proposed candidate generator alone is competitive with complex end-to-end systems from prior work.

Our sieve-based approach, detailed in \cref{algorithm:generator}, tries searches ordered from least precise to most precise until we find ontology entries that match the location mention.
Intuitively, our goal is for mentions like \textit{Austria} to match the entry  \textsc{Austria} [2782113] in GeoNames before it matches \textsc{Australia} [2077456], but still allow a typo like \textit{Australa} to match \textsc{Australia} [2077456].

\begin{algorithm}[t]
  \captionsetup{margin={-\algomargin,\algomargin}}
  \DontPrintSemicolon
  \KwIn{a location mention, $m$ \newline
        a maximum number of candidates, $k$ \newline
        the GeoNames ontology, $E$}
  \KwOut{a list of candidate entries $E_m$}
  \tcp{Index ontology}
  $I \gets \emptyset$\;
  \For{$e \in E$}{
    $name \gets \textsc{CanonicalName}(E, e)$\;
    $synonyms \gets \textsc{Synonyms}(E, e)$\;
    \For{$n \in \{name\} \cup synonyms$}{
      $I \gets I \cup \{\textsc{CreateDocument}(n, e)\}$\;
    }
  }
  \tcp{Search for candidates}
  $E_m \gets \emptyset$ \;
  \For{$t \in \{$ \textsc{Exact}, \textsc{Fuzzy}, \textsc{CharacterNgram}, \textsc{Token}, \textsc{Abbreviation}, \textsc{CountryCode} $\}$}{
    $E_m \gets \textsc{Search}(I, m, t)$\;
    \If{$E_m \ne \emptyset$}{
      break \;
    }
  }
  \tcp{Select top entries by population}
  $E_m \gets \textsc{Sort}(E_m, \textsc{key} \!=\! e \rightarrow \textsc{Population}(E, e))$\;
  \Return{} top $k$ elements of $E_m$ \;
  \caption{Candidate generator.}
  %by Lucene Search. For each entry, create a document for each name in it. Add name, abrreviation and country code into fields to do different kinds of search for the mention name. }
  \label{algorithm:generator}
\end{algorithm}

We create one document in the index for each name $n_e$ of an entry $e$ in the GeoNames ontology.
A location mention $m$ is matched to a name $n_e$ by attempting a search with each of the following matching strategies, in order:
\begin{description}[font={\bfseries\scshape}]
\item[Exact] $m$ exactly matches (ignoring whitespace) the string $n_e$ 
\item[Fuzzy] $m$ is within a 2 character Levenshtein edit distance (ignoring whitespace) of $n_e$
\item[CharacterNGram] $m$ has at least one character 3-gram overlap with $n_e$
\item[Token] $m$ has at least one token (according to the Lucene StandardAnalyzer) overlap with $n_e$
\item[Abbreviation] $m$ exactly matches the capital letters of $n_e$
\item[CountryCode] $e$ is a country and $m$ exactly matches a $e$'s country code
\end{description}
Once one of the searches has retrieved a list of matching names, we recover the ontology entry for each name, sort those ontology entries by their population in the GeoNames ontology, and return the $k$ most populous ontology entries.
This list, $E_m$ is then the input to the candidate reranker.

\subsection{Candidate Reranker}
\label{section:reranker}

Our candidate reranker is inspired by work on medical concept normalization \cite{xu-etal-2020-generate,ji2020bert}.
The reranker takes a mention, $m$, and the list of candidate entities from the candidate generator, $E_m$, encodes them with a transformer network, and uses these encoded representations to perform classification over the list to select the most probable entry.
Formally, the model prediction, $\textsc{GeoNorm}(m, E_m) = \hat{e}$, is calculated as:
\begin{align*}
    s^i &= \textsc{ToInput}(m, E_m^i) \\
    \mathbf{A}^i &= \textsc{Transformer}(s^i) \\
    \mathbf{b}^i &= \mathbf{A}_0^i \oplus \log(\textsc{Pop}(E, E_m^i))  \oplus \textsc{Type}(E, E_m^i) \\
    c^i &= (\mathbf{b}^i \mathbf{W}_{1}^T)\mathbf{W}_{2}^T \\
    \hat{\mathbf{y}} &= \softmax(c^0 \oplus \ldots \oplus c^k)
\end{align*}
where:
\begin{itemize}
    \item $E_m^i$ is the $i^{\text{th}}$ candidate entry for mention $m$
    \item $\textsc{ToInput}(m, e)$ produces a string of the form \CLS\ $m$ \SEP\ $C(E, e)$ \SEP\ $S(E, e)_1$ \SEP\ \ldots\ \SEP\ $S(E, e)_{|S(E, e)|}$ \SEP, where $C(E, e)$ is the canonical name of $e$ in the ontology, and $S(E, e)$ is the list of alternate names of $e$ in the ontology.
    \item $\textsc{Transformer}(s)$ tokenizes the string $s$ into word-pieces and produces contextualized embeddings for each of the word-pieces.
    \item $\mathbf{A}^i_0$ is the contexualized representation for the \CLS\ token of candidate entry $i$'s input string
    \item $\textsc{Pop}(E, e)$ is the population of concept $e$ in the ontology $E$
    \item $\textsc{Type}(E, e)$ is a one-hot vector identifying which of the $T$ types in the ontology $E$ the concept represents\footnote{GeoNames has $T=681$ types. For example, PPLC means \textit{capital of a political entity}. Definitions for all types (``feature codes'') are at \url{http://download.geonames.org/export/dump/featureCodes_en.txt}}
    \item $\oplus$ denotes vector concatenation
    \item $W_1 \in \mathbb{R}^{150\times {(H+1+T)}}$ and $W_2 \in \mathbb{R}^{1\times 150}$ are learned weight matrices, where $H$ is the transformer's hidden dimension
    \item $\hat{\mathbf{y}}$ is a probability distribution over the $k$ entries proposed by the candidate generator
\end{itemize}
We represent the mention text + candidate entity synonyms with the contextualized representation of the \CLS\ token, similar to applications of transformers to text classification.
We include the population feature to allow the model to learn that locations in text are more likely to refer to high population than low population places (e.g., Paris, France vs. Paris, Texas, USA), and we take the logarithm of the population under the assumption that it is more important to capture the order of magnitude (e.g., thousands vs. millions) than the exact number.
We include the type feature to allow the model to learn that locations in text are more likely to refer to some types of geographical features than others (e.g., San José, the capital of Costa Rica, vs. San José, the province).

The candidate reranker is trained with a standard classification loss:
\begin{align*}
    L_R &= \mathbf{y} \cdot log(\hat{\mathbf{y}})
\end{align*}
where $\mathbf{y} \in \mathbb{R}^{|E_m|}$ is a one-hot vector representing the correct candidate entry.

\subsection{Context Incorporation}
The text around a mention may provide clues (e.g., the context \textit{Minnesota State Patrol urges motorists to drive with caution\ldots in Becker, Clay, and Douglas} suggests that \textit{Clay} refers to Clay County, Minnesota, even though Clay County, Missouri is more populous).
Thus, we consider two approaches to incorporating context.

\paragraph{context=$c$sent}
A simple approach is to take the $c$-sentence window surrounding the mention $m$ and encode it with the the same transformer as was used to encode $m + e$.
The contextualized representation of the $c$-sentence window's \CLS\ token can then be concatenated into $\mathbf{b}$ alongside the other features.
The 512 word-piece limit on the size of the transformer input means that this approach cannot incorporate the entire document.

\paragraph{context=2stage}
To include the full document context, we take advantage of the fact (demonstrated in \cref{sec:performance_hierarchy}) that toponyms at the top of the hierarchy, like countries and states, can often be resolved precisely without context as they are less ambiguous.
We thus propose \Cref{algorithm:2-stage-resolution}, a two-stage approach to geocoding. 
Lines 3-7 are the context-free stage, where GeoNorm is first applied to all location mentions.
If the feature type of a predicted entry, $\textsc{type}(e)$, is an administrative district 1-3 (i.e., the top of the geographic hierarchy: countries, states, or counties), then the prediction is accepted.
Such predictions are converted to their administrative codes (e.g., \textit{United States} $\rightarrow$ \texttt{US}) and added to the context.
Lines 8-11 are the second stage, where the geocoding system is applied to all remaining location mentions but this time incorporating the collected context.
The context is formed by concatenating together the collected toponym codes, where for example, if Canada (CA) and Alberta (01) were found in the document as in \cref{fig:architecture}, the context string would look like ``CA || 01''.

\begin{algorithm}[t]
  \captionsetup{margin={-\algomargin,\algomargin}}
  \DontPrintSemicolon
  \KwIn{location mentions, $M$ \newline
  GeoNames ontology, $E$}
  $\hat{R} \gets \{\}$ \;
  $C \gets \emptyset$ \;
  \tcp{Resolve toponyms without context}
  \For{$m \in M$}{
    $\hat{e} \gets \textsc{GeoNorm}(m, E)$\;
    \If{$\textsc{type}(\hat{e}) \in \{\texttt{adm1},\texttt{adm2},\texttt{adm3}\}$}{
      $\hat{R}[m] \gets \hat{e}$\;
      $C \gets C \cup \{\textsc{code}(\hat{e})$\} \;
    }
  }
  \tcp{Resolve toponyms with context}
  $c \gets \texttt{"||".join}(C)$ \;
  \For{$m \in M$}{
    \If{$m \not\in \hat{R}$}{
      $\hat{R}[m] \gets \textsc{GeoNorm}(m + c, E)$\;
    }
  }
  \Return{$\hat{R}$}\;
  \caption{Two-stage toponym resolution using document-level context.}
  \label{algorithm:2-stage-resolution}
\end{algorithm}

\section{Experiments}

\subsection{Datasets}

We conduct experiments on three toponym resolution datasets.
Local Global Lexicon \citep[LGL;][]{lieberman2010geotagging} was constructed from 588 news articles from local and small U.S. news sources.
GeoWebNews \citep{gritta2019pragmatic} was constructed from 200 articles from 200 globally distributed news sites.
TR-News \citep{kamalloo2018coherent} was constructed from 118 articles from various global and local news sources.
As there are no standard publicly available splits for these datasets, we split each dataset into a train, development, and test set according to a 70\%, 10\% , and 20\% ratio.
To enable replicability, we will release these splits upon publication.
The statistics of all datasets are shown in \cref{tab:datasets}.

\begin{table}
\small
\centering
\setlength{\tabcolsep}{0.5em}
\begin{tabular}{ l r r r r r r }
\toprule
Dataset
& \multicolumn{2}{c}{Train}
& \multicolumn{2}{c}{Dev.}
& \multicolumn{2}{c}{Test} \\
\cmidrule(lr){2-3}
\cmidrule(lr){4-5}
\cmidrule(lr){6-7}
& Topo. & Art.
& Topo. & Art.
& Topo. & Art. \\ 
\midrule
LGL        & 3112 & 411 & 419 & 58 & 931 & 119 \\ 
GeoWebNews & 1641 & 140 & 281 & 20 & 477 &  40 \\ 
TR-News    &  925 &  82 &  68 & 11 & 282 &  25 \\
\bottomrule
\end{tabular}
\caption{Numbers of articles (Art.) and manually annotated toponyms (Topo.) in the train, development, and test splits of the toponym resolution corpora.}
\label{tab:datasets}
\end{table}

\subsection{Database}
\begin{figure}
\centering
    \includegraphics[width=1.0\columnwidth]{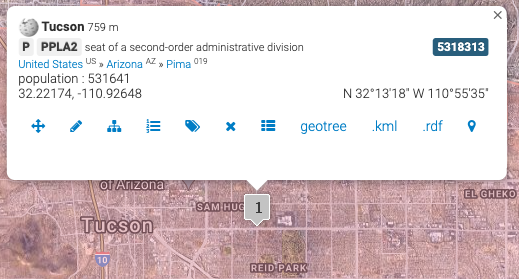}
    \caption{An entry for \textit{Tucson} in GeoNames}
    \label{fig:geonames-entry}
\end{figure}

Our datasets use GeoNames\footnote{\small\url{https://www.geonames.org/}}, a crowdsourced database of geospatial locations, with almost 7 million entries and a variety of information such as geographic coordinates (latitude and longitude), alternative names, feature type (country, city, river, mountain, etc.), population, elevation, and positions within a political geographic hierarchy.
An example entry from GeoNames is shown in \cref{fig:geonames-entry}.
%The data in GeoNames comes from multiple sources\footnote{\url{http://www.geonames.org/data-sources.html}}, such as public and open gazetteers, which can vary in quality, scope, resolution, or age \citep{ahlers2013assessment}.
%Users can edit data in a wiki-like interface.

\subsection{Evaluation Metrics}
There is not yet agreement in the field of toponym resolution on a single evaluation metric.
Therefore, we gather metrics from prior work and use all of them for evaluation.

\paragraph{Accuracy} is the number of location mentions where the system predicted the correct database entry ID, divided by the number of location mentions.
Higher is better, and a perfect model would have accuracy of 1.0.

\paragraph{Accuracy@161km} measures the fraction of system-predicted (latitude, longitude) points that were less than 161 km (100 miles) away from the human-annotated (latitude, longitude) points.
Higher is better, and a perfect model would have Accuracy@161km of 1.0.

\paragraph{Mean error distance} calculates the mean over all predictions of the distance between each system-predicted and human-annotated (latitude, longitude) point.
Lower is better, and a perfect model would have a mean error distance of 0.0.

\paragraph{Area Under the Curve} calculates the area under the curve of the distribution of geocoding error distances.
Lower is better, and a perfect model would have an area under the curve of 0.0.

\subsection{Implementation details}
We implement the candidate reranker with Lucene\footnote{\url{https://lucene.apache.org/}} v8.4.1 under Java 1.8.
When indexing GeoNames, we also index countries under their adjectival forms in Wikipedia\footnote{\url{https://en.wikipedia.org/wiki/List_of_adjectival_and_demonymic_forms_for_countries_and_nations}}.
We implement the candidate reranker with the PyTorch\footnote{\url{https://pytorch.org/}} v1.7.0 APIs in Huggingface Transformers v2.11.0 \citep{wolf-etal-2020-transformers}, using either \texttt{\small bert-base-uncased} or \texttt{\small bert-multilingual-uncased}.
We train with the Adam optimizer, a learning rate of 1e-5, a maximum sequence length of 128 tokens, and a number of epochs of 30. We explored a small number of learning rates (1e-5, 1e-6, 5e-6) and epoch numbers (10, 20, 30, 40) on the development data.
When training without context, we use one Tesla V100 GPU with 32GB memory and a batch size of 8.
When training with context, we use four Tesla V100 GPU with 32GB memory and a batch size of 32. The total number of parameters in our model is 168M and the training time is about 3 hours.

\subsection{Systems}
\label{section:systems}
We compare to a variety of geocoding systems:

\paragraph{Edinburgh}
\citet{grover2010use} introduced a rule-based extraction and disambiguation system that uses heuristics such as population count, spatial minimization, type, country, and some contextual information (containment, proximity, locality, clustering) to score, rank, and choose a candidate.

\paragraph{Mordecai}
\citet{halterman2017mordecai} introduced a generate-and-rank approach that uses Elasticsearch to generate candidates and neural networks based on word2vec \citep{word2vec} to rerank them.
Its models are trained on proprietary data.

\paragraph{CamCoder}
\citet{gritta-etal-2018-melbourne} introduced a tile-classification approach that combines a convolutional network over the target mention and 400 tokens of context with a population vector derived from location mentions in the context and populations from GeoNames.
CamCoder predicts one of 7823 tiles of the earth's surface.
See \cref{sec:appendix-camcoder} for further CamCoder details.

\paragraph{DeezyMatch}
\citet{hosseini-etal-2020-deezymatch} introduced a vector-space approach that first pre-trains an LSTM-based classifier on GeoNames taking string pairs as input, and then fine-tunes the pair classifier on the target dataset.
The trained DeezyMatch model compares mentions to database entries by generating vector representations for both and measuring their L2-norm distance or cosine similarity.

\paragraph{SAPBERT}
\citet{liu2021self} introduced a vector-space approach that pretrains a transformer network on the database using a self-alignment metric learning objective and online hard pairs mining to cluster synonyms of the same concept together and move different concepts further away.
The pre-trained SAPBERT is then fine-tuned on the target dataset.
SAPBERT was trained for the biomedical domain, but is easily retrained for other domains.
We pretrain SAPBERT on GeoNames and finetune it on the toponym resolution datasets.

% HACK: moved here for better positioning
\begin{table*}
\small
\centering
\begin{tabular}{ l c c c c c c }
\toprule
Model
& \multicolumn{2}{c}{LGL (test)}
& \multicolumn{2}{c}{GeoWebNews (test)}
& \multicolumn{2}{c}{TR-News (test)}
\\
\cmidrule(lr){2-3}
\cmidrule(lr){4-5}
\cmidrule(lr){6-7}
& R@1 & R@20
& R@1 & R@20
& R@1 & R@20
\\
\midrule
DeezyMatch \citep{hosseini-etal-2020-deezymatch}
& .172 & .538 & .262 & .671 & .206 & .702\\
SAPBERT \citep{liu2021self}
& .245 & .742 & .428 & .746 & .355 & .780 \\
GeoNorm (+gen, -rank)
& .606 & .962 & .694 & .866 & .716 & .965 \\
\bottomrule
\end{tabular}
\caption{Performance of candidate generators on the test sets. R@1 is useful for measuring the accuracy of the candidate generator when used directly as a geocoder. R@20 is useful for estimating the ceiling performance of a top-20 reranker based on that candidate generator.}
\label{tab:test_performance_generation}
\vspace{-0.25\baselineskip} %HACK
\end{table*}

\begin{table*}
\centering
\small
\setlength{\tabcolsep}{0.2em}
\begin{tabular}{l r r r r r r r r r r r r}
\toprule
& \multicolumn{4}{c}{LGL (test)}
& \multicolumn{4}{c}{GeoWebNews (test)}
& \multicolumn{4}{c}{TR-News (test)} \\
\cmidrule(lr){2-5}
\cmidrule(lr){6-9}
\cmidrule(lr){10-13}
\multicolumn{1}{c}{Model}
& Acc & A161 & Err & AUC
& Acc & A161 & Err & AUC
& Acc & A161 & Err & AUC \\
\midrule
%Lucene+Pop & & & & 0.61 & 0.33 & 16 & 0.54 & 0.67 & 0.29 & 19 & 0.60 & 0.76 & 0.21 & 12 & 0.66 \\
Edinburgh \citep{grover2010use} & .611 & - & - & - & .738 & - & - & - & .750 & - & - & - \\
CamCoder \citep{gritta-etal-2018-melbourne} & .580 & .651 & 82 & .288 & .572 & .665 & 155 & .290 & .660 & .778 & 89 & .196 \\
Mordecai \citep{halterman2017mordecai} & .322 & .375 & 926 & .594 & .291 & .333 & 1072 & .633 & .472 & .553 & 6558 & .427 \\
DeezyMatch \citep{hosseini-etal-2020-deezymatch} & .172 & .182 & 654 & .704 & .262 & .323 & 537 & .601 & .206 & .220 & 741 & .705 \\
SAPBERT \citep{liu2021self} & .245 & .260 & 566 & .630 & .428 & .499 & 357 & .446 & .355 & .362 & 595 & .568 \\
ReFinED \citep{ayoola-etal-2022-improving} & .576 & - & - & - & .658 & - & - & - & .720 & - & - & - \\
ReFinED (fine-tuned) & .786 & - & - & - & .782 & - & - & - & .858 & - & - & - \\
\midrule
GeoNorm (+gen -rank) & .606 & .685 & 119 & .263 & .694 & .774 & 92 & .194 & .716 & .812 & 95 & .169 \\
GeoNorm (+gen +rank, -context) & .761 & .785 & 59 & .167 & .788 & .834 & 61 & .131 & .798 & .816 & 89 & .154 \\
%GeoNorm & \checkmark & 1 & & .759 & .783 & 67 & .166 & -- & -- & -- & -- & -- & -- & -- & -- \\
%GeoNorm & \checkmark & 0 & & -- & -- & -- & -- & .782 & .832 & 60 & .131 & -- & -- & -- & -- \\
%GeoNorm & \checkmark & 1 & & -- & -- & -- & -- & -- & -- & -- & -- & .777 & .798 & 92 & .166 \\
%GeoNorm (+gen +rank, +context=3sent) & .759 & .783 & 67 & .166 & - & - & - & - & - & - & - & - \\
%GeoNorm (+gen +rank, +context=1sent) & - & - & - & - & .782 & .832 & 60 & .131 & - & - & - & - \\
%GeoNorm (+gen +rank, +context=3sent) & - & - & - & - & - & - & - & - & .777 & .798 & 92 & .166 \\
%GeoNorm (+gen +rank, +context=3sent) & .740 & .771 & 73 & .182 & .818 & .855 & \textbf{50} & \textbf{.113} & .805 & .812 & 97 & .158 \\
GeoNorm (+gen +rank, +context=2stage) & \textbf{.807} & \textbf{.824} & \textbf{46} & \textbf{.135} & \textbf{.828} & \textbf{.862} & \textbf{55} & \textbf{.114} & \textbf{.918} & \textbf{.933} & \textbf{34} & \textbf{.057} \\
\midrule
GeoNorm (+gen +rank, +context=2stage, +alldata) & .799 & .828 & 52 & .136 & .832 & .876 & 54 & .104 & .897 & .911 & 36 & .073 \\
\bottomrule
\end{tabular}
\caption{Performance on the test sets.
Higher is better for accuracy (Acc) and accuracy@161km (A161).
Lower is better for mean error (Err) and area under the error distances curve (AUC).
We do not report distance-based metrics for Edinburgh or ReFinED as these extraction+disambiguation systems do not make predictions for all mentions.
The best performance on each dataset+metric is in bold (excluding the final model that was trained on more data).}
\label{tab:test_performance_resolution}
\end{table*}

\paragraph{ReFinED}
\citet{ayoola-etal-2022-improving} introduced a vector-space approach for joint extraction and disambiguation of Wikipedia entities.
One transformer network generates contextualized embeddings for tokens in the text, another generates embeddings for entries in the ontology, and tokens are matched to entries by comparing dot products over embeddings.
ReFinED was trained on Wikipedia, and Wikipedia entries for place names have GeoNames IDs, so ReFinED can be used as a geocoder.
%, and ReFinED can be further finetuned on the toponym resolution datasets.

\paragraph{ReFinED (fine-tuned)}
ReFinED can also be fine-tuned, so we take the released version of ReFinED and fine-tune it for geocoding on each of the toponym datasets.

\section{Results}
\label{sec:results}
We first evaluate our context-free candidate generator, comparing it to recent context-free candidate generators.
\Cref{tab:test_performance_generation} shows that our approach outperforms approaches from prior work by large margins, both in accuracy of the top entry (R@1) and whether the correct entry is in the top 20 (R@20).

We next evaluate our complete generate-and-rank system against other geocoders.
We first perform model selection on the development set as described in \cref{section:model-selection} to select four models to run on the test set:
the candidate generator alone,
the best generate-and-rank system with no context,
and the best generate-and-rank system with context.
\Cref{tab:test_performance_resolution} shows that our proposed GeoNorm model outperforms all prior work across all toponym resolution test sets on all metrics.
Even without incorporating context, our generate-and-rank framework meets or exceeds the performance of almost all models from prior work.
The exception is ReFinED, where our context-free model outperforms ReFinED out-of-the-box, but slightly underperforms our finetuned version of ReFinED.
However, adding the novel two-stage document-level context yields large gains over the context free version of our model, and outperforms even the finetuned ReFinED.
The final row the table shows the peformance of a model trained on the combined training data from all datasets, which we release for English geocoding under the Apache License v2.0, for off-the-shelf use at  \url{https://github.com/clulab/geonorm}.

\begin{table*}
\small
\centering
\setlength{\tabcolsep}{0.2em}
\begin{tabular}{@{} c l l l l l c c c c c @{}}
\toprule
& Example & \multicolumn{4}{c}{Candidate} & \multicolumn{5}{c}{Rank} \\
\cmidrule(lr){3-6}\cmidrule(lr){7-11}
& & Name & Pop. & Type & State & RF & G & GR & GRC3 & GRCD \\
\midrule
1 & \multirow{3}{*}{\parbox{0.35\textwidth}{\textit{The educational philosophy at the Washington Latin School in \underline{Alexandria} is somewhat similar to Ahlstrom's previous endeavors.}}}
& \textbf{Alexandria} & 159467 & PPLA2 & & & & & & 1\\
& & City of Alexandria    & 139966 & ADM2 & & 1 & &\\
\\
\midrule
% 1 & \multirow{3}{*}{\parbox{0.35\textwidth}{\textit{the \underline{D.C.} Charter School Board disagreed with Ahlstrom's proposal to move the school from Northwest \underline{D.C.} to Penn Quarter.}}}
% & Washington & 689545 & PPLC & & 1 & &\\
% & & \textbf{District of Columbia}    & 552433 & ADM1 & & & & & & 1\\
% \\
% \midrule
2 & \multirow{4}{*}{\parbox{0.35\textwidth}{\textit{It was \underline{Los Angeles} police officers she attempted to blow up.}}}
& Los Angeles County & 9818605 & ADM2 & & & 1 & 2 \\
& & \textbf{Los Angeles} & 3971883 & PPLA2 & & & 2 & 1 \\
& & Los Angeles & 125430 & PPLA2 & & & 3 & 3 \\
& & Los Angeles & 4217 & PPL & & & 4 & 4 \\
\midrule
3 & \multirow{5}{*}{\parbox{0.35\textwidth}{\textit{the Minnesota State Patrol urges motorists to drive with caution as flooding continues to affect area highways. Water over the roadway is currently affecting the following areas in Becker, \underline{Clay}, and Douglas}}}
& Clay County & 221939 & & Missouri & & & 1 & & 4 \\
& & Clay County & 190865 & & Florida & & & 2 & & 3 \\
& & \textbf{Clay County} & 58999 & & Minnesota & & & 3 & & 1 \\
& & Clay County & 26890 & & Indiana & & & 4 & & 2 \\
\\
\midrule
4 & \multirow{4}{*}{\parbox{0.35\textwidth}{\textit{he writes, as do my efforts to insure \underline{New London} is a safe community.}}}
& New London County & 274055 & ADM2 & & & 1 & & 3 & 4\\
& & New London & 27179 & PPL & & & 2 & & 1 & 1 \\
& & New London & 7172 & PPL & & & 3 & & 2 & 3 \\
& & \textbf{New London} & 1882 & PPL & & & 4 & & 4 & 2 \\
\bottomrule
\end{tabular}
\caption{Examples of predictions from ReFinED (RF), our candidate generator alone (G), our generate-and-rerank system without context (GR), our system with sentence context (GRC3), and our system with 2-stage document context (GRCD).
Target location mentions are underlined.
Human annotated ontology entries are in bold.}
\label{table:error-analysis}
\end{table*}

\section{Qualitative Analysis}

\Cref{table:error-analysis} shows some qualitative analysis of errors that ReFinED and different variants of GeoNorm made.
Row 1 shows an example where ReFinED fails but GeoNorm succeeds, by more effectively using geospatial metadata such as population and feature type.
Row 2 shows an example where GeoNorm fails with a candidate generator alone but succeeds with a context-free reranker, by not relying on population alone and instead jointly considering the name, population, and feature type information (ADM2 represents a county, PPLA2 represents a city).
Row 3 shows an example where GeoNorm fails without context but succeeds with context, by taking advantage of the \textit{Minnesota} in the context to select the \textit{Clay County} that would otherwise seem implausible due to its lower population.
Finally, row 4 shows an example where our best GeoNorm model still fails.
The candidate generator includes the correct ontology entry in its top-k list, but neither the name, population, feature code, nor nearby context suggest the correct candidate.
The global context includes toponyms from the same state, allowing the model with document context to move the correct answer up from rank 4 to rank 2.
But fully addressing this issue would likely require predicting countries and states of toponyms in the text before resolving them.

\section{Limitations}
GeoNorm's candidate generator is based on information retrieval.
This is efficient but not very flexible in string matching, and when the candidate generator fails to produce the correct candidate entry, the candidate reranker also necessarily fails.
For example, as \cref{tab:test_performance_generation} shows, GeoNorm's reranker achieves only .866 recall@20 on the GeoWebNews dataset, meaning that 13.4\% of the time, the correct candidate is not in the top 20 results returned by the candidate generator.
One solution might be to replace the information retrieval based candidate generator with a neural network to provide more robust string matching, though the neural network candidate generators from prior work in \cref{tab:test_performance_generation} actually perform worse than GeoNorm's candidate generator.
Another solution may be to find smarter ways to filter the generated candidates, perhaps by building on the two-stage resolution approach to use document-level context to filter the candidates to those in appropriate countries and states.

GeoNorm is also limited by its training and evaluation data, which covers only thousands of English toponyms from news articles, while there are many millions of toponyms in many different languages across the world.
It is likely that there are regional differences in GeoNorm's accuracy that will need to be addressed by future research.

\section{Conclusion}
\label{sec:conclusion}
We propose a new toponym resolution architecture, GeoNorm, that combines the tight ontology integration of generate-and-rerank systems with the robust text encoding of deep neural networks.
GeoNorm consists of an information retrieval-based candidate generator, a BERT-based reranker that incorporates features important to toponym resolution such as population and type of location, and a novel two-stage resolution strategy that incorporates document-level context.
We evaluate our proposed architecture against prior state-of-the-art, using multiple evaluation metrics and multiple datasets.
GeoNorm achieves new state-of-the-art performance on all datasets.

\bibliography{anthology,custom}

\appendix
\setcounter{table}{0}
\renewcommand{\thetable}{\Alph{section}\arabic{table}}

\newpage

\section{Appendix}

\subsection{Performance by toponym type}
\label{sec:performance_hierarchy}

\Cref{tab:performance_hierarchy} shows that without context, GeoNorm is most precise at resolving toponyms at the top of the hierarchy, like countries and states.
\begin{table*}
\small
\centering
\setlength{\tabcolsep}{0.5em}
\begin{tabular}{ l r r r r r r r r}
\toprule
Dataset
& \multicolumn{4}{c}{Precision}
& \multicolumn{4}{c}{Recall}
\\
\cmidrule(lr){2-5}
\cmidrule(lr){6-9}
& Country & State & County & Other
& Country & State & County & Other
\\
\midrule
LGL & 0.968 & 0.806 & 0.829 & 0.745 & 0.893 & 0.915 & 0.739 & 0.763 \\
GWN & 1.000 & 0.765 & 0.778 & 0.752 & 0.966 & 0.591 & 1.000 & 0.810 \\
TR-News & 1.000 & 1.000 & 0.000 & 0.830 & 1.000 & 1.000 & 0.000 & 0.830\\ 
\bottomrule
\end{tabular}
\caption{Precision and recall of GeoNorm (without context) on three geocoding development sets.}
\label{tab:performance_hierarchy}
\end{table*}

\subsection{CamCoder details}
\label{sec:appendix-camcoder}
The original CamCoder code, when querying GeoNames to construct its input population vector from location mentions in the context, assumes it has been given canonical names for those locations.
Since canonical names are not known before locations have been resolved to entries in the ontology, we have CamCoder use mention strings instead of canonical names for querying GeoNames.

\subsection{Model selection}
\label{section:model-selection}
We performed model selection on the development sets as shown in \cref{tab:dev_performance_resolution}.
\begin{table*}
\centering
\small
\begin{tabular}{l r r r r r r r r r r r r }
\toprule
& \multicolumn{4}{c}{LGL (dev)}
& \multicolumn{4}{c}{GeoWebNews (dev)}
& \multicolumn{4}{c}{TR-News (dev)} \\
\cmidrule(lr){2-5}
\cmidrule(lr){6-9}
\cmidrule(lr){10-13}
\multicolumn{1}{c}{Model}
& Acc & A161 & Err & AUC
& Acc & A161 & Err & AUC
& Acc & A161 & Err & AUC \\
\midrule
% Edinburgh & .666 & .676 & 147 & .260 & \multicolumn{1}{l}{} & .687 & .721 & 174 & .250 & \multicolumn{1}{l}{} & .662 & .676 & 183 & .273 & \multicolumn{1}{l}{} \\
% CamCoder & .604 & .695 & 144 & .285 & \multicolumn{1}{l}{} & .509 & .712 & 174 & .293 & \multicolumn{1}{l}{} & .824 & .882 & 128 & .119 & \multicolumn{1}{l}{} \\
% Mordecai & .303 & .341 & 999 & .621 & \multicolumn{1}{l}{} & .356 & .480 & 943 & .528 & \multicolumn{1}{l}{} & .530 & .559 & 605 & .419 & \multicolumn{1}{l}{} \\
% DeezyMatch & .239 & .279 & 531 & .626 & \multicolumn{1}{l}{} & .260 & .299 & 480 & .617 & \multicolumn{1}{l}{} & .250 & .250 & 1005 & .719 & \multicolumn{1}{l}{} \\
% SAPBERT & .279 & .327 & 474 & .584 & \multicolumn{1}{l}{} & .434 & .480 & 375 & .463 & \multicolumn{1}{l}{} & .324 & .324 & 917 & .649 & \multicolumn{1}{l}{} \\
% \midrule
GeoNorm G & .594 & .671 & 201 & .289 & .644 & .858 & 73 & .165 & .677 & .735 & 187 & .242 \\

\midrule
GeoNorm GR & .802 & .819 & 64 & .141 & .865 & \underline{.925} & 39.5 & \underline{.072} & \textbf{.897} & \textbf{.912} & 64.0 & \underline{.081} \\
GeoNorm GRP & .792 & .819 & 68 & .141 & .861 & .918 & 34.7 & \underline{.072} & .868 & .882 & 65.7 & .100 \\
GeoNorm GRT & \underline{.807} & \textbf{.828} & 61 & \underline{.134} & .865 & .915 & \underline{31.9} & .073 & \textbf{.897} & \textbf{.912} & \textbf{42.7} & \textbf{.074} \\
GeoNorm GRPT & .797 & \underline{.821} & \textbf{57} & .140  & \textbf{.886} & \textbf{.940} & \textbf{29.8} & \textbf{.060} & \underline{.882} & \underline{.897} & \underline{63.5} & .090 \\
GeoNorm GRPTM  & \textbf{.814} & \textbf{.828} & \underline{60} & \textbf{.132} & \underline{.879} & .922 & 43.2 & \underline{.072} & \underline{.882} & \underline{.897} & 65.0 & .092 \\

\midrule
GeoNorm GRPTC1 & .807 & .823 & \underline{55} & .132 & .865 & .915 & 39.3 & .075 & .882 & .882 & 110 & .109  \\
GeoNorm GRPTC3 & .807 & .816 & 65 & .142 & .868 & .918 & 40.3 & .073 & .882 & .897 & 64.9 & .092 \\
GeoNorm GRPTC5 & .802 & .814 & 68 & .145 & .865 & .911 & 42.8 & .078 & .897 & .912 & 64.0 & .081 \\
GeoNorm GRPTMC1 & \underline{.816} & .831 & 62 & .133 & .872 & \textbf{.940} & \textbf{23.5} & \textbf{.057} & .882 & .897 & 64.6 & .090 \\
GeoNorm GRPTMC3 & .809 & \underline{.833} & 59 & \underline{.129} & \underline{.875} & .922 & 35.4 & .073 & \underline{.912} & \underline{.927} & \underline{40.6} & \underline{.063} \\
GeoNorm GRPTMC5 & .807 & .823 & 61 & .137 & .872 & \textbf{.940} & \underline{29.4} & \underline{.060} & .868 & .882 & 72.6 & .103 \\
GeoNorm GRPTMCD & \textbf{.885} & \textbf{.897} & \textbf{29} & \textbf{.079} & \textbf{.879} & \underline{.925} & 31.0 & .065 & \textbf{.971} & \textbf{.985} & \textbf{6.8} & \textbf{.010}  \\ 
\bottomrule
\end{tabular}
\caption{Performance on the development sets.
Higher is better for accuracy (Acc) and accuracy@161km (A161).
Lower is better for mean error (Err) and area under the error distances curve (AUC).
The top score in each group is in bold, the second best score is underlined.
Model features are indicated by the string of characters: G means the candidate generator was applied, R means a reranker was applied, P means the reranker included the population feature, T means the reranker included the type feature, M means the reranker was fine-tuned from \texttt{\small bert-multilingual-uncased} instead of \texttt{\small bert-base-uncased}, C1/C3/C5 means the reranker included 1/3/5 sentences of context, and CD means the reranker included the two-stage document-level context algorithm.
%Edinburgh made no predictions for 24, 16, and 0 toponyms in LGL, GeoWebNews, and TR-News, respectively.
%Since it is impossible to calculate distance-based metrics without coordinates, we skip those toponyms, and as a result overestimate the distance-based metrics for Edinburgh.
}
\label{tab:dev_performance_resolution}
\end{table*}
All GeoNorm models that included a reranker (R) outperformed the candidate generator (G) alone.
We explored the population (P) and type (T) features in models without context, and found that they helped slightly on LGL and GeoWebNews but hurt slightly on TR-News.
For models with context, rerankers fine-tuned from \texttt{\small bert-multilingual-uncased} (M) slightly outperformed models fined-tuned from \texttt{\small bert-base-uncased}.
Adding sentence level context (C1/C3/C5) to the rerankers helped on TR-News, but did not help on LGL or GeoWebNews.
Applying the two-stage algorithm for document-level context led to large gains on LGL and TR-News, but did not help on GeoWebNews.

We thus selected the following models for evaluation: GeoNorm G, GeoNorm GRPT, and GeoNorm GRPTMCD.

\subsection{Artifact intended use and coverage}
\label{sec:appendix-intended-use}
The intended use of \texttt{\small bert-base-uncased} and \texttt{\small bert-multilingual-uncased} is to be ``fine-tuned on tasks that use the whole sentence''\footnote{\url{https://huggingface.co/bert-base-uncased}}.
We have used them for that purpose when encoding the context, but also for the related task of encoding place names, which are usually short phrases.
These artifacts are trained on English books and English Wikipedia and released under an Apache 2.0 license which is compatible with our use.

The intended use of our geocoding model is matching English place names in text to the GeoNames ontology.
Though GeoNames covers millions of place names, our evaluation corpora cover only English news articles, and thus the performance we report is only predictive of performance in that domain.

\end{document}